\documentclass[12pt]{article}

\usepackage{amsmath}
\usepackage{amsfonts}
\usepackage{amssymb}
\usepackage{makecell}
\usepackage{booktabs}
\usepackage{caption}
\usepackage{float}
\usepackage[margin=1in, nohead]{geometry}
\usepackage{graphicx}
\usepackage[colorlinks,linkcolor=blue,citecolor={blue!50!black},urlcolor={blue!80!black},bookmarks=false,hypertexnames=true]{hyperref}
\usepackage{lscape}
\usepackage[sc]{mathpazo}
\usepackage{mathtools}
\usepackage{apacite}
\usepackage{pdflscape}
\usepackage{rotating}
\usepackage{setspace}
\usepackage[flushleft]{threeparttable}
\usepackage{verbatim}
\usepackage{xcolor}
\usepackage{subcaption}
\usepackage{subfiles}
\usepackage{placeins}
\usepackage{amsthm}

\theoremstyle{definition}

\theoremstyle{remark}

\usepackage{natbib}

\begin{document}
\title{On Bob Dylan: A Computational Perspective}


\author{\begin{tabular}{c}
     Prashant Garg\footnote{Imperial College London, Department of Economics and Public Policy. Email: prashant.garg@imperial.ac.uk}
 \end{tabular}}
\date{\today}
\maketitle
\vspace{-1cm}
\begin{center}
Preliminary - click \href{https://drive.google.com/file/d/1lnXV02XAz1Z2rQ1teGuO2F_Tm4PhaspJ/view?usp=sharing}{\textbf{here}} for the latest version
\end{center}

\begin{abstract}
\noindent  Cass Sunstein’s essay \textit{On Bob Dylan} describes Dylan’s “dishabituating” style—a constant refusal to conform to expectation and a penchant for reinventing his musical and lyrical identity. In this paper, I extend Sunstein’s observations through a large-scale computational analysis of Dylan’s lyrics from 1962 to 2012. Using o3-mini-high (a large language model), I extract concept-to-concept relationships from the lyrics and construct directed knowledge graphs that capture Dylan’s thematic structure. I then quantify shifts in sentiment, metaphorical expression, thematic diversity, and network complexity over time. The results indicate that Dylan’s lyrics increasingly rely on metaphor, display an evolving sentiment profile, and exhibit heightened dishabituation—measured here as a growing variance in the network centrality of key concepts. I also find that references to movement, protest, and mythic imagery fluctuate in ways that align with well-known phases of Dylan’s career, reflecting the dynamic and unpredictable quality of his art. These findings not only deepen our empirical understanding of Sunstein’s thesis but also introduce a novel computational method for analyzing an artist’s evolution—offering broader applicability to the study of cultural and creative change.

\bigskip

  \smallskip
\end{abstract}

\setstretch{1.25}

\clearpage
\section{Introduction}

Bob Dylan’s career spans more than five decades and covers an extensive range of stylistic shifts—from early folk protest music to electric rock, country-tinged albums, and gospel-infused work. His ability to reinvent himself, both musically and lyrically, has long been recognized as a defining feature of his artistry. \citet{sunstein2022bob} describes Dylan’s capacity to \textit{dishabituate}, highlighting his persistent drive to disrupt conventional thought through continual reinvention. 

This idea aligns with psychological research on habituation and dishabituation, which suggests that human beings become desensitized to repeated stimuli over time, unless those stimuli are altered or recontextualized in ways that restore their impact \citep{sharot2025look, rankin2009habituation, groves1970habituation}. \citet{sharot2024look} explores how breaking routine patterns can restore attentional engagement and cognitive flexibility. Dylan’s work exemplifies this principle through his frequent stylistic shifts and resistance to genre conventions, ensuring that his audience is continually surprised. His approach aligns with the idea that dishabituation—whether in perception, cognition, or artistic expression—is central to maintaining freshness and engagement.

BBuilding on these insights, this study employs large language models (LLMs) and network analysis to quantitatively examine Dylan’s lyrical evolution. In doing so, the work not only investigates Dylan’s shifting use of literal versus metaphorical language, sentiment, and thematic prevalence but also demonstrates a new method to understand the personal evolution of an artist—a methodological approach that can be extended to other cultural figures. 

Specifically, I investigate whether Dylan’s evolving use of literal and metaphorical language, shifts in sentiment, fluctuations in thematic prevalence, and variations in conceptual centrality collectively reflect a deliberate process of dishabituation. To this end, I construct a directed concept-to-concept network from Dylan’s lyrics (1962–2012), where each node represents a concept (e.g., “love,” “road,” “heaven,” “war”), and each edge carries annotations on sentiment, expression style (literal or metaphorical), and relationship type.

\section{Data and Methods}

I assembled a corpus of Bob Dylan’s studio album lyrics from 1962 to 2012, excluding live-only recordings, purely instrumental tracks, and unreleased outtakes. Track titles, album identifiers, and release years were compiled into a structured dataset that formed the basis for further analysis.

\paragraph{LLM-Based Data Extraction.}  
I used o3-mini-high, a large language model (LLM), to process each song in segmented chunks and extract pairs of related concepts.\footnote{OpenAI o3-mini, released on January 31, 2025, is among the most advanced and cost-effective reasoning models available. It offers robust STEM capabilities, supports function calling and structured outputs, and achieves high performance with low latency. For more details, see \url{https://openai.com/index/openai-o3-mini/}} For each extracted edge, the LLM annotated sentiment (positive, negative, or neutral), classified the relationship as literal or metaphorical, and assigned a broad relationship type (correlational or causal). The output was stored in a structured format, linking each concept pair with its corresponding lyric excerpt. Carefully designed prompts ensured that the LLM adhered to a predefined JSON schema, thereby facilitating efficient and consistent processing of the large volume of data. This approach adopts the framework of \cite{gargfetzer2024_causalclaims}, which was originally developed to extract knowledge graphs from academic papers, and adapts it for processing song lyrics.

\paragraph{Network Construction and Analysis.}  
From these extracted edges, I constructed a global directed network in which nodes represent normalized concepts (merging synonyms such as “car” and “automobile”) and edges capture the relationships identified by the LLM. I computed standard network measures—such as degree and eigenvector centrality—at both the node and song levels. In particular, I used the variance in eigenvector centrality within each song as a proxy for dishabituation; a higher variance indicates that a song blends frequently referenced (mainstream) concepts with those that are rarely mentioned (peripheral).

\paragraph{Thematic Analysis.}  
To examine the evolution of lyrical themes over time, I developed thematic dictionaries for nine categories: Protest/Political, Mythic/Biblical, Movement/Travel, Emotional, Mortality, Love/Romance, Nature, Violence, and Time. These dictionaries were derived from the node list produced by the LLM by querying another instance of the model (o3-mini-high) to cluster nodes with more than two connections into thematic groups. For example, the Movement/Travel dictionary includes terms such as “road,” “train,” “highway,” “journey,” and “wander.” Edges were flagged if either the source or sink concept matched a term from a thematic dictionary, allowing me to track the annual prevalence of each theme over time.

\section{Data and Methods}

I assembled a corpus of Bob Dylan’s studio album lyrics from 1962 to 2012, excluding live-only recordings, purely instrumental tracks, and unreleased outtakes. Track titles, album identifiers, and release years were compiled into a structured dataset that formed the basis for further analysis.

\paragraph{LLM-Based Data Extraction}  
To extract concept-to-concept relationships from the lyrics, I employed a large language model (o3-mini-high) using a comprehensive prompt. In this prompt, the model was assigned the role of an expert assistant tasked with analyzing a song’s title, album, album year, and complete lyrics. It was instructed to identify all distinct relationships between concepts—whether explicit or symbolic—and to output a structured JSON object for each relationship. 

Each JSON object includes a claim in the format “A -> B” (where A is the source concept and B is the sink concept), along with key annotations: the source and sink concepts, the overall sentiment (categorized as positive, negative, or neutral), the effect size (small, medium, large, or NA), the relationship type (e.g., causal or correlational), and a classification of whether the relationship is expressed literally or metaphorically (with an option for ambiguous or NA).

Additionally, the model recorded a snippet of the lyric, a confidence level for the extraction, and categorized both source and sink nodes into predefined types (such as person, object, emotion, place, abstract, event, or other). If no clear relationships were detected, the model returned an empty array. Carefully designed prompts ensured that the LLM adhered to a predefined JSON schema. This extraction framework builds on the methodology described in \cite{gargfetzer2024_causalclaims}, originally developed to extract knowledge graphs from academic texts, and adapts it for processing song lyrics.

\paragraph{Embedding Standardization.}
Following extraction, I standardized the free-text node labels using an embedding-based matching approach. I initialized the OpenAIEmbeddings model with the configuration \texttt{model="text-embedding-3-large", dimensions=3072} to compute high-dimensional vector embeddings for each node’s description. A controlled vocabulary with standardized 1–2 word terms and brief definitions was then used to merge semantically similar labels (e.g., “car” and “automobile”) and assign each node to broader categories. Appendix Section~\ref{controlled_vocab} describes the generation of this controlled vocabulary. For each node, its embedding was compared (via cosine similarity) to the embeddings of vocabulary entries, and the term with the highest similarity was assigned as the standardized label. This process enhanced the consistency and interpretability of the network by ensuring that variants of the same concept were unified under a single term.

\paragraph{Network Construction and Analysis.}  
Using the structured output from the extraction and standardization steps, I constructed a global directed network in which nodes represent normalized concepts and edges capture the relationships identified by the LLM. Standard network measures—such as degree, eigenvector centrality, PageRank, betweenness, and closeness—were computed at both the node and song levels. In particular, I used the variance in eigenvector centrality within each song as a proxy for dishabituation; a higher variance indicates that a song blends frequently referenced (mainstream) concepts with those that are rarely mentioned (peripheral).

\paragraph{Thematic Analysis.}  
To examine the evolution of lyrical themes over time, I developed thematic dictionaries for nine categories: Protest/Political, Mythic/Biblical, Movement/Travel, Emotional, Mortality, Love/Romance, Nature, Violence, and Time. These dictionaries were derived from the node list produced by the LLM by querying another instance of the model (o3-mini-high) to cluster nodes with more than two connections into thematic groups. For example, the Movement/Travel dictionary includes terms such as “road,” “train,” “highway,” “journey,” and “wander.” Edges were flagged if either the source or sink concept matched a term from a thematic dictionary, and the annual prevalence of each theme was computed for subsequent analysis.

\section{Results}

\paragraph{Network Visualization.}  

To explore the evolution of conceptual connectivity in Dylan’s lyrics, I constructed cumulative networks for two distinct periods: the first 25 years (1962–1986) and the second 25 years (1987–2012). In these networks, each node represents a unique concept extracted from the lyrics, and an edge denotes a relationship identified by the LLM. Node degrees were computed to assess centrality, and the top 20 nodes by degree were visualized for each period.

\begin{figure}[H]
    \centering
    \caption{Networks of Dylan's Lyrical Concepts}
    \label{fig:cumulative_networks}
    
    \begin{subfigure}{\textwidth}
        \centering
        \caption{\small \textbf{Panel A (1962--1986)}}
        \includegraphics[width=0.75\textwidth]{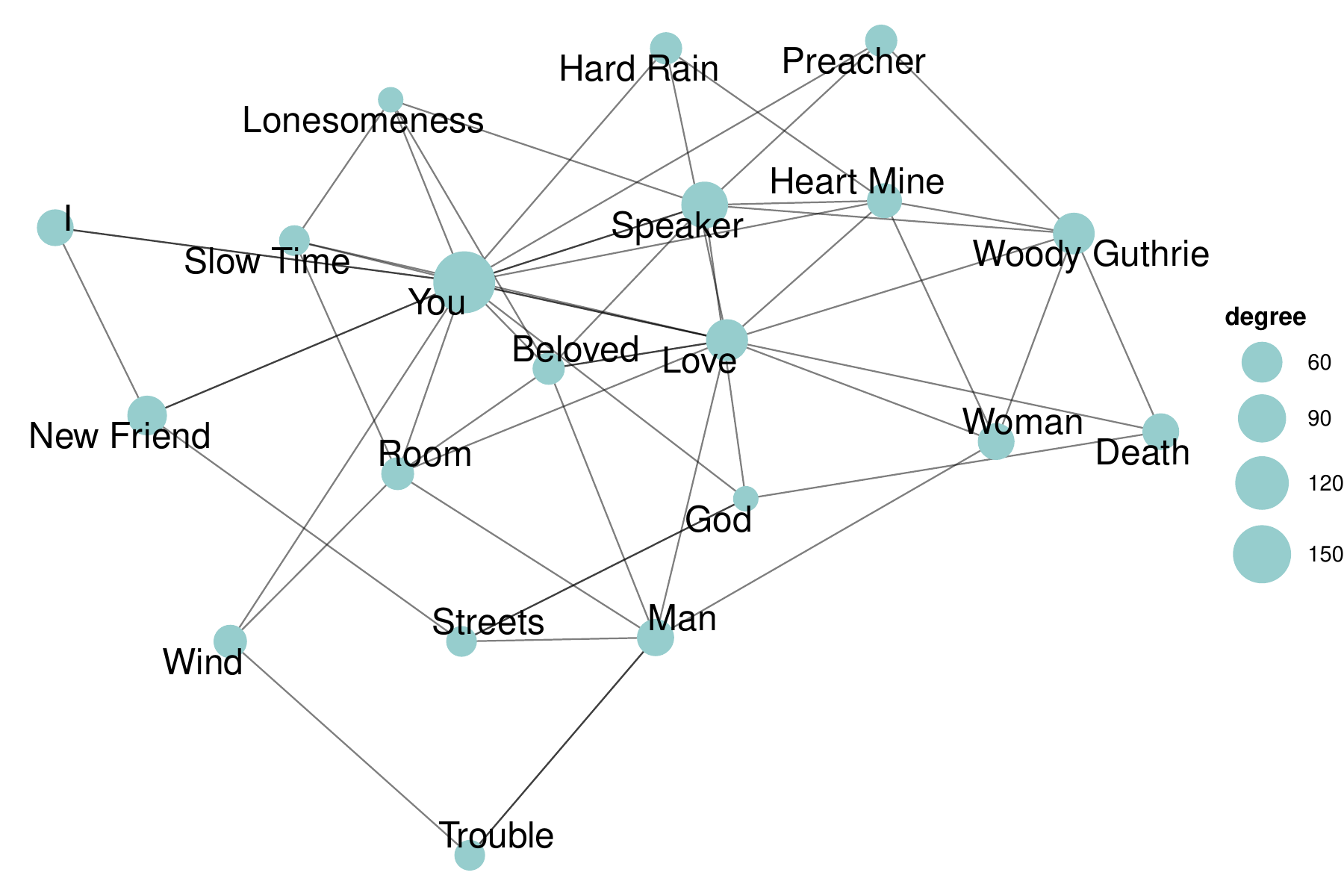}
    \end{subfigure}
    
    \begin{subfigure}{\textwidth}
        \centering
        \caption{\small \textbf{Panel B (1987--2012)}}
        \includegraphics[width=0.75\textwidth]{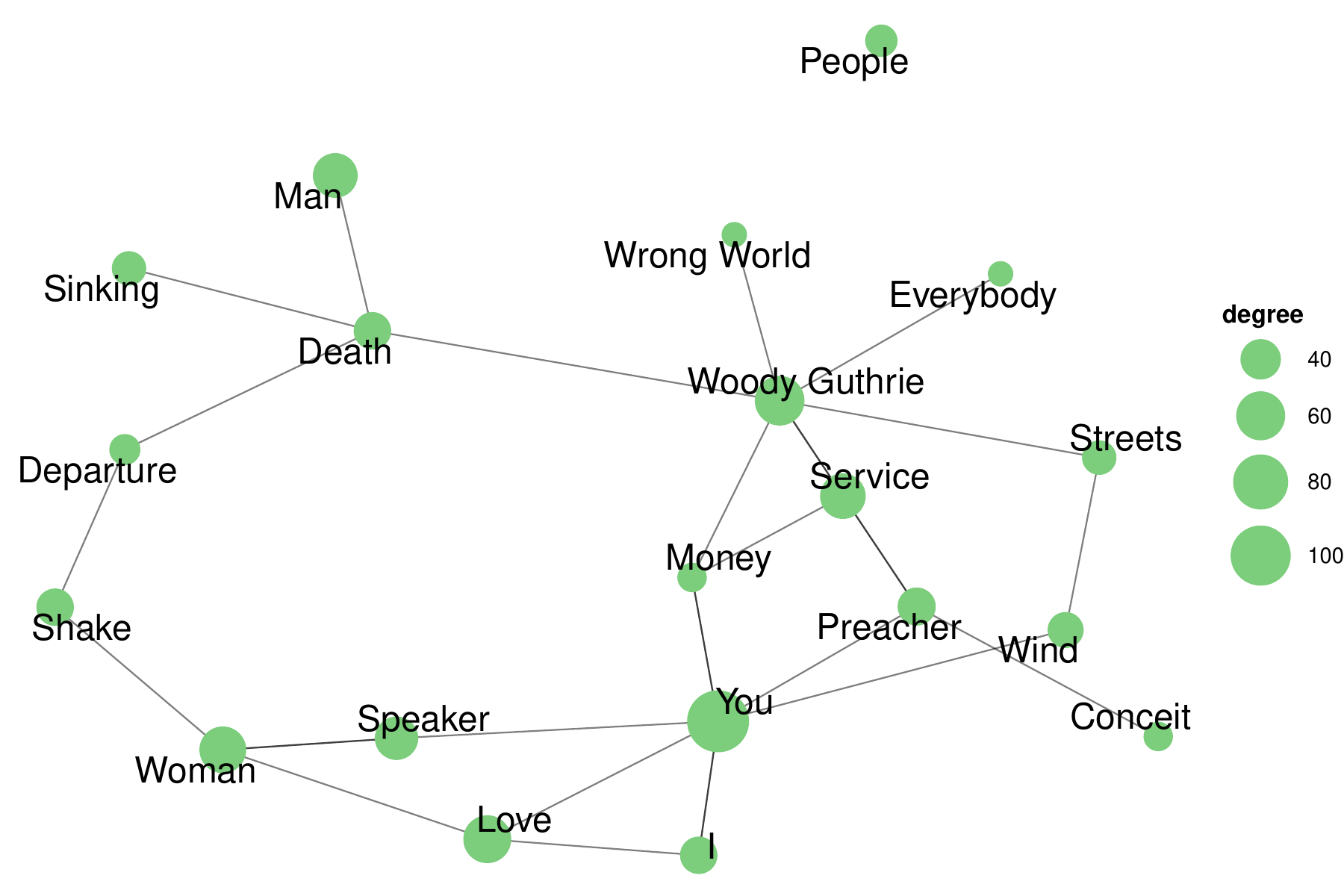}
    \end{subfigure}
    \caption*{\small \textbf{Note:} \textit{Panel A (1962--1986)} displays a network dominated by personal address, poetic abstraction, and historical influences, whereas \textit{Panel B (1987--2012)} reveals an evolution toward themes of transience, broader societal engagement, and economic critique.} 
\end{figure}

In the 1962–1986 network (Panel A of Figure~\ref{fig:cumulative_networks}), highly central nodes include “You,” “Love,” “Man,” “Beloved,” “Woody Guthrie,” “Wind,” “Hard Rain,” “Preacher,” “God,” and “Death.” This pattern reflects an early emphasis on personal address, poetic abstraction, historical influence, and existential themes—consistent with Dylan’s folk revival and protest-song era. In contrast, the 1987–2012 network (Panel B) retains “You” as the most central node but sees the emergence of “Departure,” “Sinking,” “Service,” and “Money” as prominent concepts. These changes indicate a shift toward themes of transience, broader social commentary, and material critique, aligning with Dylan’s later work. (See Figure~\ref{fig:cumulative_networks} below.)

\paragraph{Theme Prevalence and Node-Type Transitions.}  
To gain insight into the structural and emotional properties of Dylan’s lyrical network, I first classified the extracted concepts into a set of node types. For example, the \textit{person} category includes references to individuals (e.g., “I,” “man,” “woman”), while \textit{emotion} covers feelings such as love or grief. The \textit{abstract} category comprises conceptual ideas (e.g., “truth,” “memory”), \textit{object} includes tangible items (e.g., “train,” “flower”), \textit{place} denotes locations, \textit{event} captures occurrences (e.g., “kiss,” “hard rain,” “departure”), and \textit{other} collects miscellaneous labels. I then aggregated edges by their source and sink node types and computed both the total count and average sentiment (coded as –1 for negative, 1 for positive, and 0 for neutral) for each pairing.

Figure~\ref{fig:alluvial_node_sentiment} presents an alluvial diagram of these transitions. The width of each flow represents the number of edges linking a given source node type (on the left) to a sink node type (on the right), and the fill color reflects the mean sentiment (darker red indicates more negative sentiment). For instance, the \textit{person} category emerges as the most frequent source (over 3,350 occurrences), and the \textit{abstract} category is the most common sink (exceeding 4,280 edges). In contrast, flows such as from \textit{event} to \textit{other} exhibit an average sentiment near –0.5, indicating a more negative emotional tone. These variations provide a window into the emotional dynamics underlying Dylan’s conceptual connections.

\paragraph{Theme Prevalence Over Time.}  
To capture how core themes evolved across decades, I applied thematic dictionaries for nine categories—Protest/Political, Mythic/Biblical, Movement/Travel, Emotional, Mortality, Love/Romance, Nature, Violence, and Time—to flag lyrical edges whose concepts matched these lists. The faceted plot in Figure~\ref{fig:theme_prevalence} displays, for each thematic category, the percentage of edges (as a proportion of all extracted edges) associated with that theme in each year. The x-axis denotes the year, and the LOESS-smoothed curves (on the y-axis, expressed as a percentage) highlight the underlying trends.

For example, the "Protest/Political" facet, which tracks terms related to political activism, government, revolution, and social justice, exhibits a prominent peak in the early-to-mid 1960s—consistent with Dylan’s celebrated protest era—followed by a steep decline. In contrast, the "Mythic/Biblical" facet (covering terms such as Cain, Abel, Angels, Heaven, and Hell) shows a marked increase during the late 1970s and early 1980s, aligning with Dylan’s born-again period. The "Movement/Travel" facet remains consistently high, with peaks in the mid-1960s and a notable reemergence in the 1990s, reflecting Dylan’s enduring fascination with rootlessness. Similar patterns emerge for the remaining themes, each echoing aspects of Dylan’s evolving lyrical concerns.

\paragraph{Literal vs. Metaphorical Expression.}  
To quantify the evolution of Dylan’s lyrical style, I asked the large language model to classify each extracted concept-to-concept edge as either “literal,” “metaphorical,” or “ambiguous.” Out of all classified edges, 3,306 were labeled as literal and 5,646 as metaphorical, while only 163 were ambiguous (and were excluded from further analysis). For example, literal edges include straightforward relationships such as “God → you” (from “May God bless and keep you always”), “Mr. Tambourine Man → song” (from “Hey, Mr. Tambourine Man, play a song for me”), or “Johnny → medicine” (from “Johnny's in the basement mixing up the medicine”). In contrast, metaphorical edges include imaginative pairings such as “ladder → stars” (from “May you build a ladder to the stars and climb on every rung”), “evening's empire → sand” (from “Though I know that evening's empire has returned into sand”), and “magic, swirling ship → trip” (from “Take me on a trip upon your magic, swirling ship”).

\begin{figure}[H]
    \centering
    \caption{Proportion of Literal versus Metaphorical Edges by Decade}
    \includegraphics[width=\textwidth]{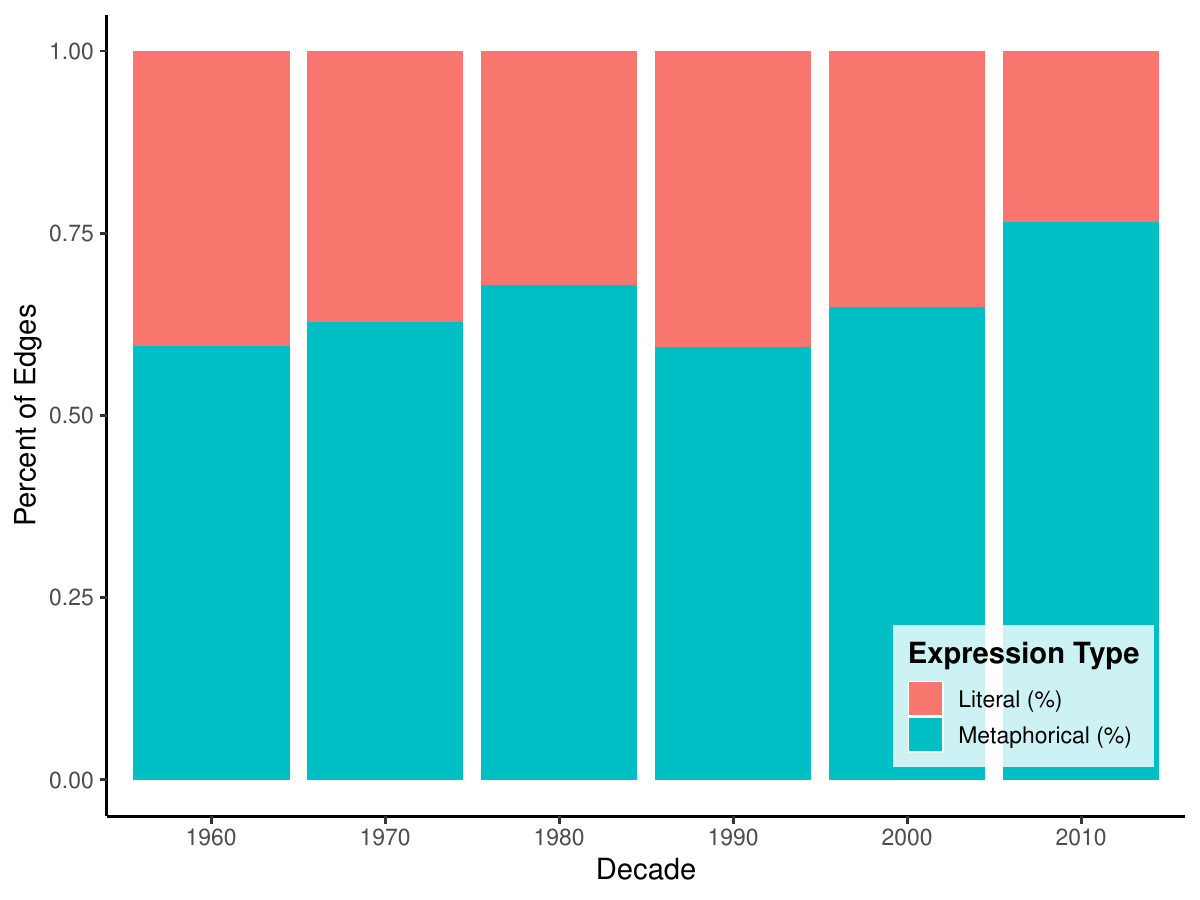}
    \caption*{\small \textbf{Note:} Each bar represents the percentage of edges (excluding ambiguous cases) that are classified as literal or metaphorical for a given decade. Here, an "edge" denotes a directed conceptual relationship extracted from Dylan’s lyrics. The plot shows a clear decline in literal expression and a corresponding increase in metaphorical expression over time, supporting the view that Dylan’s later work relies more on symbolic language.}
    \label{fig:literal_meta_example}
\end{figure}

Figure~\ref{fig:literal_meta_example} displays the proportion of literal and metaphorical edges (excluding ambiguous cases) for each decade. The data show that literal edges accounted for roughly 40–42\% of relationships in the 1960s and 1970s but steadily declined to about 23.5\% by the 2010s. Conversely, the proportion of metaphorical edges increased from about 58.5\% in the 1960s to over 76\% in the 2010s. Additionally, I find that metaphorical edges tend to be associated with more negative sentiment and larger effect sizes than literal ones. These findings support the view that Dylan’s later work increasingly relies on symbolic, emotionally charged language—a key element of dishabituation.


\paragraph{Dishabituation via Centrality Variance.}  
To quantify the extent to which Dylan’s songs blend widely recognized (high-centrality) concepts with more obscure (low-centrality) ones—a phenomenon I interpret as dishabituation—I computed the variance in eigenvector centralities for the set of nodes present in each song. In the network, each node represents a concept extracted from the lyrics, and an edge links two concepts based on an LLM-identified relationship. A higher variance in eigenvector centrality indicates a greater disparity between frequently referenced (mainstream) and rarely mentioned (peripheral) concepts, suggesting a more eclectic mix of ideas.

\begin{figure}[H]
    \centering
    \caption{Normalized Variance in Eigenvector Centralities Across Decades (1960--2010)}
    \includegraphics[width=0.75\textwidth]{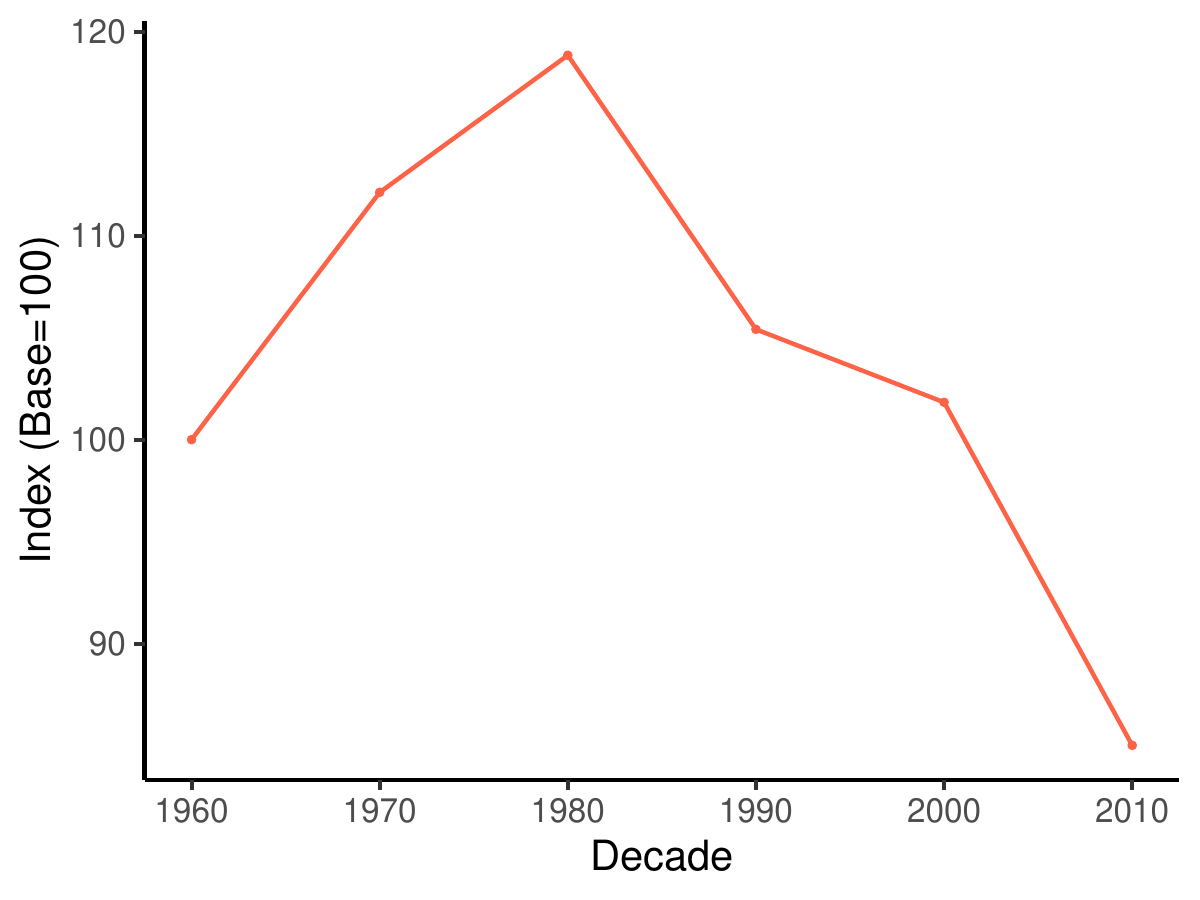}
    \caption*{\small \textbf{Note:} The plot shows the mean variance in eigenvector centralities computed for songs within each decade, normalized to an index of 100 for the 1960s. A higher index indicates a greater disparity between high- and low-centrality nodes, reflecting a stronger mix of mainstream and peripheral concepts. The index increases from 100 (1960s) to approximately 119 in the 1980s, then declines to around 85 by the 2010s, suggesting that the highest level of dishabituation occurs mid-career, with later work exhibiting a reduction in this heterogeneity.}
    \label{fig:variance_centrality_example}
\end{figure}

I aggregated the variance measures over 10-year periods by first calculating the mean variance in eigenvector centrality for songs within each decade, then normalizing these values by setting the 1960s as the baseline (index = 100). As shown in Figure~\ref{fig:variance_centrality_example}, the normalized mean variance increases from 100 in the 1960s to approximately 119 in the 1980s—indicating that during this period, Dylan’s lyrics exhibited the greatest mix of common and uncommon references. Interestingly, the variance declines in the 1990s and 2000s, reaching an index of about 85 in the 2010s. This suggests that while Dylan’s mid-career work may have maximized dishabituation through a broad conceptual mix, his later work appears more thematically cohesive or constrained.

\section{Discussion}
\begin{figure}[H]
    \centering
    \caption{Alluvial Diagram of Node-Type Transitions Colored by Mean Sentiment}
    \includegraphics[width=\textwidth]{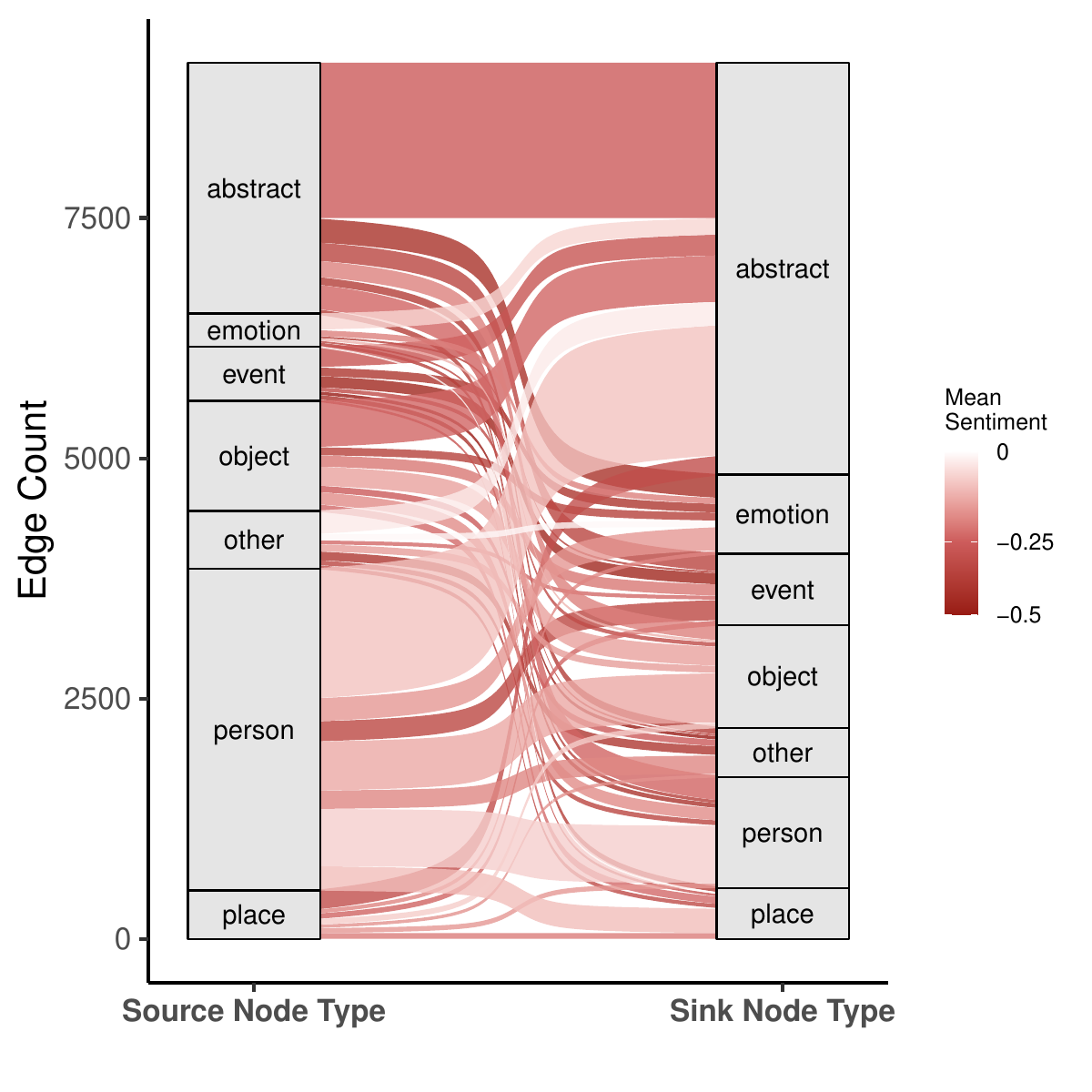}
    \caption*{\small \textbf{Note:} This diagram visualizes transitions between source and sink node-type categories in the lyrical network. The width of each flow represents the number of concept-to-concept edges linking the respective categories, while the fill color indicates the mean sentiment (with values ranging from –0.5 for highly negative to 0 for neutral). For example, the flow from \textit{person} to \textit{abstract} (with a count of 1,353 edges and mean sentiment near –0.05) suggests that many personal references are connected to abstract ideas, typically with a nearly neutral tone.}
    \label{fig:alluvial_node_sentiment}
\end{figure}

The analyses reveal that Dylan’s lyrical network evolves significantly over time in both its thematic content and the emotional interactions among its concepts. Figure \ref{fig:alluvial_node_sentiment} shows that personal references (e.g., the prevalent \textit{person} category) often converge toward abstract ideas, while certain transitions (e.g., from \textit{event} to \textit{other}) carry a more negative tone. The thematic prevalence analysis indicates that protest/political ideas were dominant in the early 1960s, followed by a mid-career surge in mythic/biblical references and a persistent emphasis on movement motifs throughout his career. In parallel, the analysis of literal versus metaphorical expression reveals a clear shift: literal relationships declined from about 40\% in the 1960s and 1970s to under 25\% by the 2010s, while metaphorical language increased correspondingly. Moreover, metaphorical edges tend to carry more negative sentiment and exhibit larger effect sizes. Moreover, the measurement of dishabituation via centrality variance shows that the disparity between mainstream (high-centrality) and peripheral (low-centrality) concepts peaked in the 1980s—suggesting that Dylan’s lyrics were most eclectic and disruptive during this period—before declining in later decades, possibly indicating a more cohesive thematic approach.

The quantitative indicators in the analysis—ranging from the evolving balance of literal and metaphorical expression to shifts in thematic prevalence and network centrality variance—strongly support Sunstein’s portrayal of Dylan as an artist who continuously disrupts conventional thought. Notably, the rising reliance on metaphor, the fluctuating prominence of themes such as protest and mythic imagery, and the peak in centrality variance during the 1980s all point to periods of heightened dishabituation in Dylan’s work. 

Moreover, the approach presents a new method to understand the personal evolution of an artist by mapping the complex interplay of lyrical concepts over time. Although the method has limitations—such as reliance on LLM interpretations and a focus solely on textual content—it offers a promising framework that can be extended to study other cultural icons and their creative transformations.

Overall, the study provides a systematic portrait of Dylan’s evolving lyrical landscape and reinforces the notion that his art remains a powerful force in challenging routine. Future research could extend these methods to other artists or incorporate additional data layers—such as live performance analysis, interviews, or audience reception studies—to offer a more comprehensive understanding of how creative disruption manifests in music.

\section{Figures}

\begin{figure}[H]
    \centering
    \caption{Thematic Prevalence Over Time}
    \includegraphics[width=\textwidth]{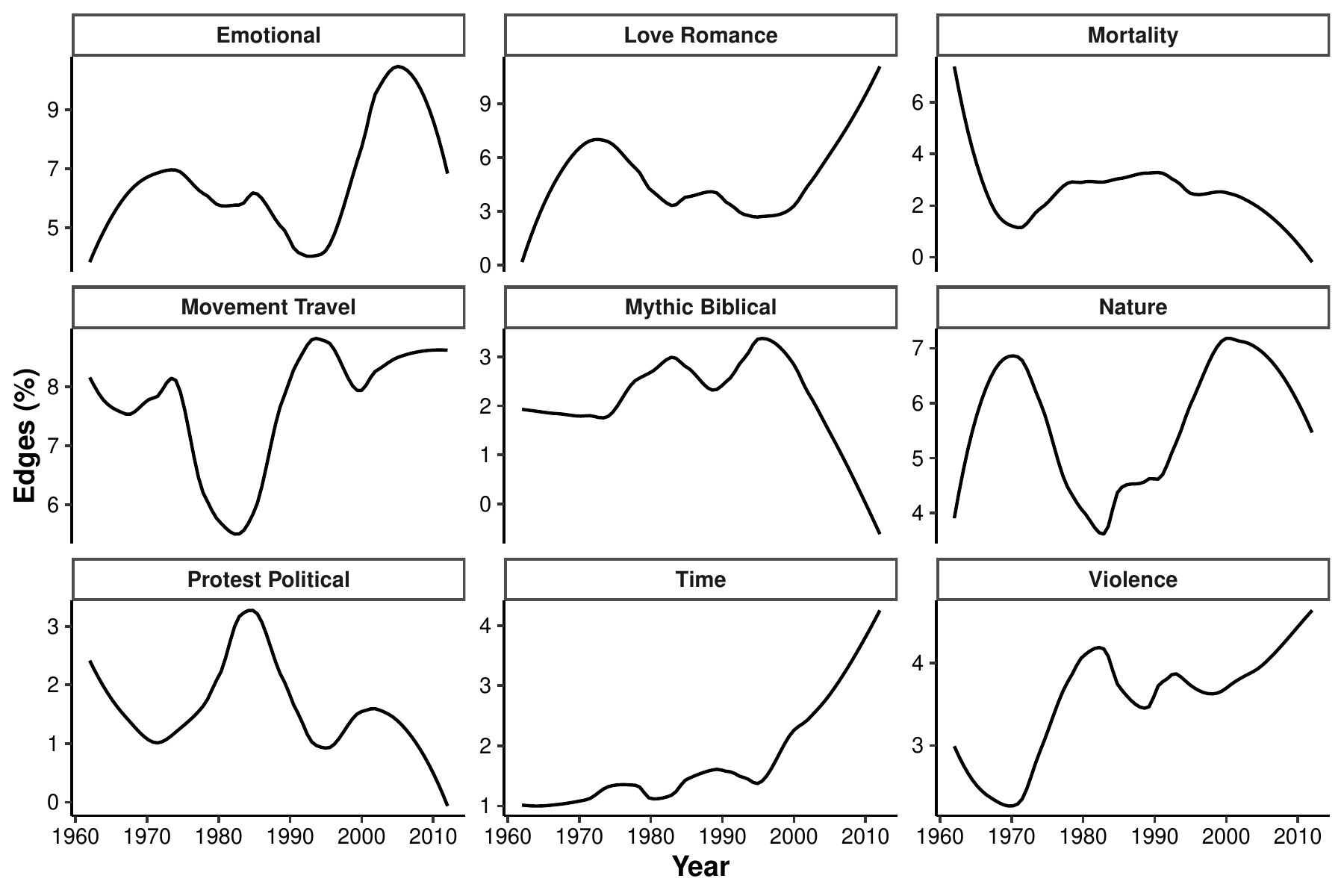}
    \caption*{\small \textbf{Note: }Each panel shows the LOESS-smoothed trend of the percentage of lyrical edges that match a specific thematic category, as defined by the thematic dictionaries. For example, the "Protest/Political" panel indicates a peak in the early-to-mid 1960s followed by a decline, while the "Mythic/Biblical" panel shows increased references during the late 1970s and early 1980s.}
    \label{fig:theme_prevalence}
\end{figure}

\appendix

\section{Additional Details on Data and Thematic Dictionaries}\label{controlled_vocab}

In the extraction process, the large language model processed each song’s title, album information, and full lyrics to extract pairs of related concepts. Each extracted edge was annotated with information regarding whether the relationship was literal or metaphorical and with its sentiment (positive, negative, or neutral). This process yielded thousands of edges spanning Dylan’s discography, all stored in a structured format for further analysis.

To ensure consistency in our network, I applied a controlled vocabulary strategy. First, I compiled a list of nodes (i.e., the extracted concept labels) and filtered out those with very few connections. I then used the LLM to generate a JSON-formatted dictionary containing standardized 1–2 word terms along with brief definitions. This allowed me to merge semantically similar labels (e.g., “car” and “automobile”) and assign each normalized node to broader categories such as person, object, place, and emotion. For a similar vector-embedding approach in a different context, see \cite{fetzer2024ai} and \cite{gargfetzer2024_causalclaims}.

A similar procedure was used to create thematic dictionaries. I focused on nodes with more than two connections and submitted them to an LLM (chatgpt o3-mini-high) with a prompt designed to cluster these nodes into thematic groups. The model classified the nodes into nine categories: Protest/Political, Mythic/Biblical, Movement/Travel, Emotional, Mortality, Love/Romance, Nature, Violence, and Time. Table~\ref{tab:theme_dictionaries} below presents sample terms for each thematic category.

After generating these dictionaries, I flagged each edge in the dataset if its source or sink concept matched any term from the appropriate dictionary. I then computed, for each year, the percentage of edges flagged for each theme, which enabled the faceted thematic analysis presented in Figure~\ref{fig:theme_prevalence} of the main text.

\begin{table}[H]
\centering
\caption{Example Thematic Dictionaries}
\begin{tabular}{p{4cm} p{11cm}}
\toprule
\textbf{Thematic Category} & \textbf{Sample Terms} \\
\midrule
Protest/Political & War, Government, Politics, Oppression, Revolution, Justice, Bomb, Civil Rights, MLK, Freedom, Russians, Senator, Authorities \\
Mythic/Biblical   & Cain, Abel, Samaritan, Noah, Ezekiel, Ophelia, Einstein, Casanova, Ezra Pound, Angels, Heaven, Hell, Sin, Salvation, Divine, Bible \\
Movement/Travel   & Leaving, Escape, Sinking, Departure, Hobo, Road, Old Road, Streets, Running, Movement, Train, Freight Train, Travel, Wandering, Journey, Car, Highway, Wiggle \\
Emotional         & Sadness, Pain, Joy, Despair, Regret, Happiness, Anger, Fear, Lonely, Grief, Sorrow, Love, Mercy, Hurt, Hatred, Negativity, Doubt \\
Mortality         & Death, Dying, Funeral, Grave, Bones, Ghost, Dead, Deaths, Afterlife, Coffin \\
Love/Romance      & Love, Beloved, Heart, Kiss, Lover, Lovers, Affection, Intimacy, Romance, True Love, Marriage, Sweetheart \\
Nature            & Rain, Hard Rain, Storm, Sky, River, Mountains, Sun, Moon, Stars, Flowers, Thunder, Snow, Wind, Clouds, Sea, Earth \\
Violence          & Violence, Killing, Gun, Hate, Crime, Fighting, Stoning, Punishment, War, Assault, Blood \\
Time              & Time, Future, Past, Hour, Clock, Later, Fast, Impermanence, Eternity, Aging \\
\bottomrule
\end{tabular}
\label{tab:theme_dictionaries}
\end{table}

\end{document}